\documentclass{article}

\usepackage[table,dvipsnames]{xcolor}
\usepackage{colortbl}

\usepackage[preprint]{corl_2025} %

\usepackage{amssymb}
\usepackage{amsmath}
\usepackage{wrapfig}
\usepackage{graphicx}
\usepackage{algorithm}
\usepackage{algorithmic}
\usepackage{subcaption}
\usepackage{booktabs}
\usepackage{siunitx}
\usepackage{microtype}
\usepackage[capitalise]{cleveref}

\usepackage{amsmath,amsfonts,bm}

\def\1{\bm{1}}

\DeclareMathAlphabet{\mathsfit}{\encodingdefault}{\sfdefault}{m}{sl}
\SetMathAlphabet{\mathsfit}{bold}{\encodingdefault}{\sfdefault}{bx}{n}

\def\gA{{\mathcal{A}}}

\def\gS{{\mathcal{S}}}

\def\gW{{\mathcal{W}}}

\usepackage{titlesec}

\crefname{equation}{Eq.}{Eqs.}
\crefname{figure}{Fig.}{Figs.}
\crefname{section}{Sec.}{Sec.}
\crefname{appendix}{App.}{App.}
\crefname{table}{Tab.}{Tabs.}
\crefname{algorithm}{Algo}{Algo}
\crefname{thm}{Thm}{Thm}
\Crefname{thm}{Thm}{Thm}
\crefname{prop}{Prop}{Prop}

\usepackage{etoolbox}

\usepackage{etoolbox}
\AtBeginEnvironment{align}{\setlength{\abovedisplayskip}{5pt}\setlength{\belowdisplayskip}{5pt}}

\title{Divide, Discover, Deploy: Factorized Skill Learning with Symmetry and Style Priors}

\author{
  Rafael~Cathomen\\
  ETH Zurich\\
  \texttt{\small{carafael@ethz.ch}} \\
  \And
  Mayank Mittal \\
  ETH Zurich \& NVIDIA \\
  \texttt{\small mittalma@ethz.ch} \\
  \And
  Marin Vlastelica\\
  ETH Zurich \\
  \texttt{\small mvlastelica@ethz.ch} \\
  \And
  Marco Hutter \\
  ETH Zurich \\
  \texttt{\small mahutter@ethz.ch} \\
}

\begin{document}
\maketitle

\newcommand{\statespace}[0]{\mathcal{S}}
\newcommand{\observationspace}[0]{\mathcal{O}}
\newcommand{\actionspace}[0]{\mathcal{A}}
\newcommand{\trajectoryspace}[0]{\mathcal{T}}
\newcommand{\distributionspace}[0]{\mathcal{P}}
\newcommand{\goalspace}[0]{\mathcal{G}}
\newcommand{\skillspace}[0]{\mathcal{Z}}
\newcommand{\symmetryspace}[0]{\mathcal{I}}
\newcommand{\mdp}[0]{\mathcal{M}}

\newcommand{\transitionoperator}[0]{A}
\newcommand{\entropy}[0]{\mathcal{H}}

\NewDocumentCommand{\transitionprobability}{om}{%
  \IfValueTF{#1}{%
    \mathcal{T}(#1 \mid #2) %
  }{%
    \mathcal{T} %
  }%
}
\newcommand{\prior}[0]{p}   

\NewDocumentCommand{\observationprobability}{om}{%
  \IfValueTF{#1}{%
    \mathcal{E}(#1 \mid #2)
  }{%
    \mathcal{E}
  }%
}

\newcommand{\goalfromstate}[0]{g}   
\newcommand{\staterepr}[0]{\phi}   
\newcommand{\reprsf}[0]{\psi}   
\newcommand{\discriminator}[0]{q}   
\newcommand{\rewardfunction}[0]{r}   
\newcommand{\occupancymeasure}[0]{\rho}   
\newcommand{\policy}[0]{\pi_{\theta}}   
\newcommand{\valuefunction}[0]{V}
\newcommand{\qfunction}[0]{Q}
\newcommand{\advantage}[0]{A}
\newcommand{\return}[0]{R}
\newcommand{\objective}[0]{J}
\newcommand{\KL}[2]{D_{\text{KL}}(#1 \Vert #2)}
\newcommand{\wasserstein}[0]{\mathcal{W}}
\newcommand{\mirrorskill}[0]{M_z}
\newcommand{\mirrorstate}[0]{M_s}
\newcommand{\mirroraction}[0]{M_a}
\newcommand{\normalize}[0]{\texttt{normalize}}

\let\originalmiddle=\middle
\def\middle#1{\mathrel{}\originalmiddle#1\mathrel{}}

\newcommand{\state}[0]{\boldsymbol{s}}   
\newcommand{\goal}[0]{\boldsymbol{g}}   
\newcommand{\skill}[0]{\boldsymbol{z}}   
\newcommand{\statedistribution}[0]{\boldsymbol{\mu}}   
\newcommand{\action}[0]{\boldsymbol{a}}   
\newcommand{\observation}[0]{\boldsymbol{o}}   
\newcommand{\hidden}[0]{\boldsymbol{h}}   

\newcommand{\skillvar}[0]{\boldsymbol{Z}}   
\newcommand{\statevar}[0]{\boldsymbol{S}}   
\newcommand{\actionvar}[0]{\boldsymbol{A}}   

\newcommand{\reward}[0]{r}   
\newcommand{\factorweight}[0]{\lambda}   
\newcommand{\discountfactor}[0]{\gamma}   
\newcommand{\learningrate}[0]{\alpha}   
\newcommand{\symmetry}[0]{i}   

\newcommand{\trajectory}[0]{\tau}   
\newcommand{\replaybuffer}[0]{\mathcal{R}}   
\newcommand{\data}[0]{\mathcal{D}}

\newif\ifcomments
\commentstrue
\newcommand{\red}[1]{\textcolor{red}{#1}}
\newcommand{\TODO}[1]{\ifcomments\red{(TODO: #1)}\else \fi}
\newcommand{\UNCLEAR}[1]{\red{UNCLEAR: #1}}
\newcommand{\rafael}[1]{\ifcomments\textcolor{blue}{\bf RC says: #1}\else \fi}
\newcommand{\mayank}[1]{\ifcomments \textcolor{orange}{\bf MM says: #1}\else \fi}
\newcommand{\marin}[1]{\ifcomments \textcolor{red}{\bf MV says: #1}\else \fi}
\newcommand{\marco}[1]{\ifcomments \textcolor{teal}{\bf MH says: #1}\else \fi}
\newcommand{\rebuttal}[1]{\textcolor{blue}{#1}}

\newcommand{\etal}{\textit{et al}. }
\newcommand{\ie}{\textit{i}.\textit{e}., }
\newcommand{\eg}{\textit{e}.\textit{g}. }

\begin{abstract}

Unsupervised Skill Discovery (USD) allows agents to autonomously learn diverse behaviors without task-specific rewards. While recent USD methods have shown promise, their application to real-world robotics remains underexplored.
In this paper, we propose a modular USD framework to address the challenges in the safety, interpretability, and deployability of the learned skills.
Our approach employs user-defined factorization of the state space to learn disentangled skill representations. It assigns different skill discovery algorithms to each factor based on the desired intrinsic reward function.
To encourage structured morphology-aware skills, we introduce symmetry-based inductive biases tailored to individual factors. We also incorporate a style factor and regularization penalties to promote safe and robust behaviors.
We evaluate our framework in simulation using a quadrupedal robot and demonstrate zero-shot transfer of the learned skills to real hardware. Our results show that factorization and symmetry lead to the discovery of structured human-interpretable behaviors, while the style factor and penalties enhance safety and diversity. Additionally, we show that the learned skills can be used for downstream tasks and perform on par with oracle policies trained with hand-crafted rewards.
For code and videos, please check: \href{https://leggedrobotics.github.io/d3-skill-discovery}{https://leggedrobotics.github.io/d3-skill-discovery/}.

\end{abstract}


\section{Introduction}

Reinforcement learning (RL) has achieved remarkable success across a range of real-world robotics applications~\citep{perceptiveLocomotionMiki2022, Kaufmann2023rldrone, openai2019solvingrubikscuberobot}.
However, these successes typically depend on carefully prespecified reward functions. 
Designing such rewards for a large number of tasks demands significant engineering effort and often becomes increasingly complex as task difficulty grows.
Unsupervised Skill Discovery (USD) seeks to address these challenges by training agents to autonomously acquire a diverse repertoire of behaviors, or \textit{skills}, without relying on handcrafted rewards. These skills can then be reused or fine-tuned to solve downstream tasks more efficiently.

\begin{figure}[thb]
    \centering
    \includegraphics[width=\linewidth]{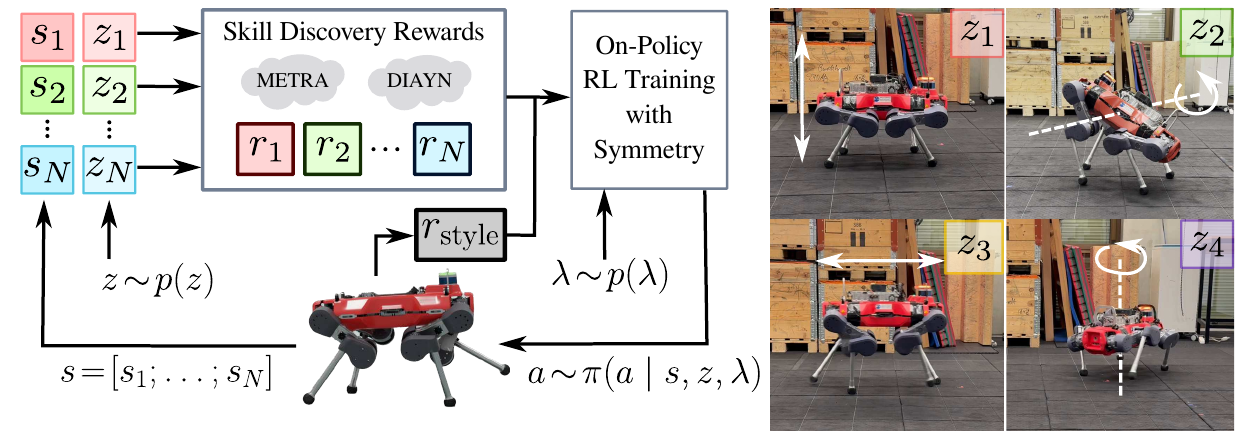}
    \vspace*{-18pt}
    \caption{
    \textbf{Approach overview.} The agent’s state $\state$ is factorized by the user into $N$ components, each paired with a latent skill $z_i$ and an intrinsic reward $r_i$, selected from METRA or DIAYN objectives. An extrinsic reward $r_{\text{style}}$ promotes safe behaviors. The factor weights \(\lambda\) allow the agent to prioritize certain factors during training. The policy $\pi$ is trained using on-policy RL with symmetry augmentation to discover structured, morphology-aware skills.
    The resulting skills are interpretable, robust, and can be commanded by a human operator.}
    \vspace*{-1em}
    \label{fig:main-figure}
\end{figure}

\looseness=-1
Current USD approaches use an intrinsic reward function to generate training signals to acquire task-agnostic behaviors. These intrinsic rewards are typically formulated as different variants of mutual information (MI) between the agent's state $\state$ and its latent skill representation $\skill$.
For instance, DIAYN~\citep{eysenbach_diversity_2018} optimizes a variational lower bound on the MI, while METRA~\cite{park_metra_2024} uses a Wasserstein variant of the MI. Furthermore, \citet{hu_disentangled_2024} show that factorizing $(\state, \skill)$ facilitates more interpretable and controllable skills. Despite these advances, most USD research remains confined to simulation, with limited demonstrations on real-world robotic systems.

A core limitation in USD lies in the exclusive reliance on intrinsic rewards: while these encourage exploration and behavioral diversity, they offer no feedback on whether the learned behaviors are safe, stable, or physically feasible on real hardware. As a result, the behaviors learned by many USD approaches tend to be overly aggressive or unsafe.
Although some works~\citep{emergentoffdads, atanassov_constrained_2024} have demonstrated unsupervised skill deployment on hardware, they usually focus on constrained or simplified scenarios. Efforts to improve safety in USD~\citep{kim_safety-aware_2023, kim_sdodont_2024} have also made progress, but often trade off between safety, skill diversity, and scalability. Overcoming these challenges is essential for advancing USD from an exploratory paradigm to a practical tool for developing robotic systems.

\looseness=-1
In this work, we present a factorized skill discovery framework that selectively applies USD algorithms across different state dimensions defined by the user (\cref{fig:main-figure}). The core idea is that the desired form of diversity often depends on the specific subset of the state space and the chosen USD algorithm. For instance, in our experiments, we observe that METRA excels at improving the state coverage on unbounded state factors such as planar position, while DIAYN is better suited for bounded state factors, such as the robot's orientation, where continuous drift is impossible.
Our design effectively takes advantage of these individual benefits while also exploiting the symmetry in the robot’s morphology. By extending this symmetry to the skill space, the framework encourages the discovery of more structured skills. To address the critical issue of deployability, we propose two additional mechanisms. First, we introduce an additional \emph{style} factor, which is an extrinsic signal that shapes the agent's behavior toward safe and stable actions. Second, we develop a skill weighting mechanism that facilitates the handling of conflicting skills and allows their balanced adjustment during deployment. Using this framework, we demonstrate the discovery of diverse quadrupedal skills that are learned entirely in simulation and can be safely deployed on real hardware. %

\section{Related Work}

\looseness=-1
\textbf{Unsupervised Skill Discovery.}
The goal of USD is to extract task-agnostic behaviors from intrinsic rewards. 
\citet{eysenbach_diversity_2018} maximize the lower bound on MI between skills and states via a learned discriminator, while \citet{sharma_dynamics-aware_2020} add transition dynamics to encourage more kinetic skills.
Optimistic exploration through discriminator ensembles~\citep{strouse_learning_2022_optimistic} further enhances the state coverage. 
This is complementary to the problem of exploration, where the goal is to maximize coverage, often regularized by task reward~\citep{burda2018explorationrandomnetworkdistillation,chen2017ucbexplorationqensembles,lee2019efficient,osband2013more,osband2016deep}, or maximize information gain~\citep{sukhija2024maxinforl}.
An alternate line of work replaces the MI objective with a Wasserstein dependency measure (WDM). 
METRA~\cite{park_metra_2024} and its variants~\cite{park2022lipschitzconstrainedunsupervisedskilldiscovery, park2023controllabilityawareunsupervisedskilldiscovery, rho_language_2024} maximize the directed distance in a learned latent space, resulting in highly dynamic state-covering skills.
To increase interpretability of learned skills, DUSDi~\citep{hu_disentangled_2024} factorizes the state and skill spaces and applies DIAYN per factor with an entanglement penalty. Subsequent work~\citep{wang2024skildunsupervisedskilldiscovery} extends this by using inter-factor dependency graphs to discover interaction-focused skills.
Our proposed framework generalizes the factorization idea in DUSDi by allowing different USD objectives per state factor, letting each dimension exploit the most suitable notion of diversity.
Additionally, we inject robot morphology-based symmetry priors \rebuttal{\citep{Apraez_2025, Su_2024, mittal2024symmetryconsiderationslearningtask}} into the latent skills and introduce factor weights to coordinate potentially conflicting skills.

\textbf{Deployment of USD.} Most USD studies remain mainly in simulation; only a few consider real robots.
\citet{kim_sdodont_2024} bias METRA with labeled desirable and undesirable trajectories, while \citet{atanassov_constrained_2024} combine a norm-matching objective with hand-crafted rewards to transfer discovered locomotion skills to a quadruped.
\citet{cheng2024learning_domino_real} utilize DOMiNO \citep{zahavy2023discoveringpoliciesdominodiversity_domino} to learn diverse solutions for navigation tasks, while still relying on explicit task rewards.
Further efforts~\citep{vlastelicaoffline,kolevdual} have been made in constructing offline task-regularized USD algorithms by leveraging the Fenchel duality.
\citet{emergentoffdads} show that basic locomotion skills can be learned directly on hardware with an off-policy version of DADS. However, the learned skills are relatively simple (planar locomotion) and contain undesirable motion artifacts. Unlike prior work relying on supervision or task rewards, we deploy intrinsically learned skills by combining a style factor, global regularization, and per-factor weighting, which balances safety with skill diversity.

\section{Background}
\label{sec:background}

\textbf{Symmetric Factored MDP.}
In this work, we consider a symmetric factored MDP. %
A factored MDP~\cite{hu_disentangled_2024} is defined as the tuple \( \mdp(\gS, \gA, \transitionprobability{}, R) \), where the state space \( \gS = \gS_1 \times  \cdots \times \gS_N \) is factorized into \(N\) factors. Each state \(\state \in \statespace\) consists of \(N\) state factors: \(\state = [\state_1;\dots;\state_N], \state_i \in \statespace_i \). The action space and transition kernel are denoted by \( \actionspace \) and \( \transitionprobability{}: \statespace \times \actionspace \to \Delta(\statespace) \) respectively, where $\Delta(\cdot)$ denotes the probability simplex. The goal of USD is to learn a skill-conditioned policy \( \policy: \statespace \times \skillspace \to \Delta(\actionspace)  \) that results in diverse, useful, and distinguishable behaviors (\ie \emph{skills}).
This is typically achieved by maximizing a certain information objective, such as the mutual information (MI) between states and latent skills. 
Following the factored MDP, the skill space \( \skillspace = \skillspace_1 \times  \cdots \times \skillspace_N \) is also factorized, with the disentangled skill component $\skill_i$ only affecting the state factor $\state_i$. The skills are sampled from a prior distribution \( \skill \sim \prior(\skill) = \Pi_{i=1}^N p(\skill_i)\). The MI objective results in a reward function $R: \statespace \times \skillspace \times \actionspace \to \mathbb{R}$, which can be maximized using standard RL~\cite{schulman2017proximalpolicyoptimizationalgorithms,haarnoja2018soft}.

Intuitively, a symmetric MDP~\cite{ravindran2001symm} means the dynamics and rewards are preserved under a set of transformations over the state and action spaces, such as a left-right reflection. An MDP has a $K$-fold symmetry if a set of $K$ distinct transformations exist under which the transition dynamics is equivariant and the reward model is invariant. Extending this definition to USD, the transformations also need to be defined over the skill space. Let the functions $\mirrorstate^k: \statespace \to \statespace$, $\mirroraction^k: \actionspace \to \actionspace$ and $\mirrorskill^k: \skillspace \to \skillspace$ define the $k$-th transformation functions for states, actions and skills, respectively. The MDP $\mdp$ is symmetric if $\forall k \in {1, \dots, K}$, $\state, \state' \in \statespace$, $\action \in \actionspace$ and $z \in \skillspace$,
the transition model $\transitionprobability[\state']{\state,\action} = \transitionprobability[\mirrorstate^k(\state')]{\mirrorstate^k(\state),\mirroraction^k(\action)}$ is equivariant, and the reward model $R(\state,\action, \skill) = R(\mirrorstate^k(\state),\mirroraction^k(\action), \mirrorskill^k(\skill))$ and the skill prior $\prior(\skill) = \prior(\mirrorskill^k(\skill))$ are invariant.
It is important to note that the mirror functions $\mirrorstate^k$ and $\mirroraction^k$ are determined solely based on the transition model (\ie the robot's morphology), while the mirror function for the skills $\mirrorskill^k$ must be defined in a way that the symmetry condition holds. These choices are discussed in \cref{sec:method}.

\looseness=-1
\textbf{MI-based USD Rewards.} Our framework utilizes two algorithms: DIAYN~\cite{eysenbach_diversity_2018} and METRA~\cite{park_metra_2024}. DIAYN maximizes MI, \( I(\statevar; \skillvar) \triangleq \KL{p(\state, \skill)}{p(\state) p(\skill)} \), using a learned discriminator \( \discriminator_{\phi}(\skill | \state) \) that approximates the posterior $p(\skill | \state)$, yielding the reward \( \reward_{\text{DIAYN}}(\state, \skill) = \log \discriminator_{\phi}(\skill | \state) - \log p(\skill)\).
METRA replaces MI with WDM: \( I_\gW(\statevar; \skillvar) \triangleq \gW(p(\state, \skill),p(\state) p(\skill)) \) under a temporal distance metric. It trains $\staterepr (\state)$ and uses the reward \(\reward_{\text{METRA}}(\state, \skill, \state') = (\staterepr(\state') - \staterepr(\state))^\top \skill \) to align latent state transitions with the skill vector.
Additional details are in~\cref{sec:app_usd}.

\section{Method}
\label{sec:method}

\begin{figure}[htbp]
  \centering
  \begin{minipage}[c]{0.39\textwidth}
    \centering
    \includegraphics[width=\linewidth]{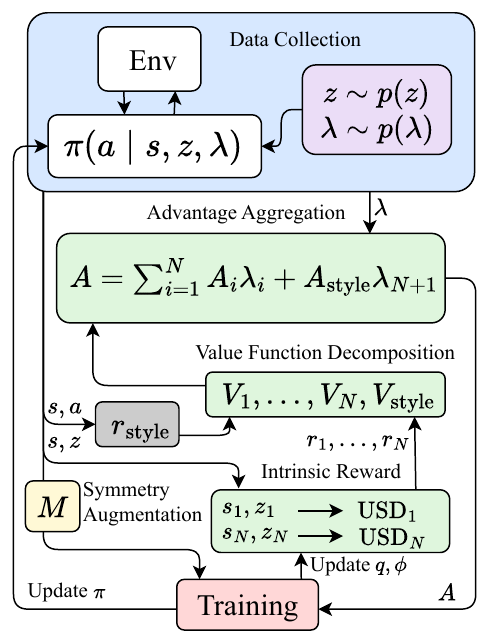 }
  \end{minipage}%
  \hfill
  \begin{minipage}[c]{0.60\textwidth}
    \small
    \centering
    \captionsetup{skip=0pt}
    \begin{algorithm}[H]
      \small 
      \caption*{\textbf{Algorithm 1:} Skill Discovery Procedure}
      \begin{algorithmic}
        \STATE Factorize state space \( \statespace = \statespace_1 \times  \cdots \times \statespace_N \)
        \STATE Factorize skill space \( \skillspace = \skillspace_1 \times  \cdots \times \skillspace_N \)
        \STATE Define skill prior $p(\skill)$ and weighting prior $p(\lambda)$
        \STATE Define style reward \( r_\text{style} \) and regularization reward \( r_\text{reg} \) 
        \STATE Define symmetry mirroring functions \( M = (\mirrorstate, \mirroraction, \mirrorskill) \)
        \STATE Select USD algorithm per factor \(\text{USD}_i \in \{\text{METRA}, \text{DIAYN}\}\)
        \\\hrulefill
        \STATE \textbf{Initialize} $\policy$, $\{\valuefunction_i\}_{i=1}^N$, $\valuefunction_\text{style}$
        \WHILE{not converged}
          \STATE Sample skill $\skill \sim p(\skill)$ and weights $\lambda \sim p(\lambda)$  every $k$ steps
          \STATE Sample action $\action \sim \policy (\action | \state, \skill, \lambda)$
          \STATE Collect on-policy samples $(\state,\state',\action,\skill,\lambda, r)$
          \STATE Augment the dataset by mirroring each sample 
          \STATE Update all $\text{USD}_i$ and $\valuefunction_i$ with the augmented samples
          \STATE Compute advantages \(A_i\) for each factorized value function
          \STATE Compute weighted sum of advantages \(A = \sum_{i=1}^{N+1} \lambda_i A_i\)
          \STATE Update $\policy$ based on $A$ using any on-policy RL algorithm
        \ENDWHILE
      \end{algorithmic}
    \end{algorithm}
  \end{minipage}
  \caption{\textbf{Proposed algorithm for skill discovery.} The agent \(\policy\), conditioned on a sampled skill \(\skill\) and factor weights \(\lambda \), collects transitions and receives a total reward combining per-factor intrinsic rewards and a style reward. The transitions are then augmented via symmetry-based mirroring, after which the intrinsic reward models, factorized value functions, and policy are updated using on-policy RL.}\label{fig:main_method}
 \vspace*{-1.5em}
\end{figure}

Our approach builds on the idea of factorizing the latent skill vector $\skill$ to create independent and disentangled skill components. Building on the notion of factored MDPs, we extend the framework from~\citet{hu_disentangled_2024} to support different intrinsic objectives and include a \emph{style} objective to promote deployable behaviors. This flexibility enables behavior-specific inductive biases by applying the most suitable USD algorithm per factor.
To further enhance control and coordination between skill components, we introduce a scalar weight for each factor, which allows prioritizing specific components during training and effectively resolving conflicts between simultaneously learned skills.
More formally, the objective for the skill-conditioned policy $\policy$ is to maximize
\begin{equation}
    \mathcal{J}(\theta) = \sum_{i=1}^N \lambda_i I_{\text{USD}_i}(\statevar_i, \skillvar_i) + \lambda_{N+1} J_\text{style}(\statevar, \actionvar),
\end{equation}
where \( \lambda = [\lambda_i]_{i=1}^{N+1} \) is the factor weighting vector, which assigns relative importance to individual objectives. The per-factor objective $I_{\text{USD}_i}$ depends on the selected USD algorithm for that factor. In this work, we consider this objective based on DIAYN with disentanglement penalty~\cite{hu_disentangled_2024,eysenbach_diversity_2018} or METRA~\cite{park_metra_2024}. The objective $J_\text{style}$ includes extrinsic rewards for neutral skills, such as standing stationary. In the remainder of the section, we provide further details on these individual components. %

\textbf{Factor Weighting.}
When we disentangle the skill space through factorization, each skill dimension (or factor) is intended to control a distinct and ideally independent aspect of the agent's behavior. However, in practice, state dependencies between factors can lead to behavioral conflicts. For example, skill factors for standing still and moving forward cannot be executed simultaneously without interference. This issue leads to poorer coverage of the learned skill factors~\cite{hu_disentangled_2024}.
To manage potential conflicts between skill factors, we introduce \textit{factor weights} \( \lambda \in \mathbb{R}^{N+1} \), where each \( \lambda_i \geq 0 \) and \( \|\lambda\|_2 = 1 \). These weights modulate the relative importance of each intrinsic reward, arising from $I_{\text{USD}_i}$, as well as any extrinsic rewards. By conditioning the policy \( \policy \) on \( \lambda \), the agent can dynamically prioritize the skill factors.
During training, we sample 
$\lambda$ by normalizing a vector of i.i.d. positive values from a truncated Gaussian, ensuring the norm constraint is satisfied.

\textbf{Style Factor and Regularization Penalties.}
To improve the deployability of unsupervised skills, prior work~\cite{atanassov_constrained_2024} incorporates two types of extrinsic rewards, one for smoothness and one for aesthetic behavior. However, applying both rewards uniformly can overconstrain skill discovery and limit diversity. To address this, we separate these signals. 
We introduce an additional factor that provides a neutral ``style" reward that depends on the robot’s configuration. For a quadruped, this reward may encourage the robot to maintain a stable posture, such as standing still.
Treating this as a separate factor allows the agent to learn a safe fallback skill and benefit from a soft inductive bias during learning. Since the style factor is included in the policy input, its influence can be modulated dynamically via the weighting mechanism, enabling the agent to stay near safe behaviors when needed, while still exploring meaningfully under intrinsic objectives.

We apply global regularization penalties to reduce joint torques and velocities and enforce physical constraints such as joint limits. Unlike the style factor, these are applied uniformly across all skills and are not part of the policy input. They promote safety and support hardware deployment.

The resulting reward is defined as: $
    \reward_{\text{}}(\state,  \action, \skill) = \sum_{i=1}^N \lambda_i   \reward_{\text{USD}_i}(\state_i, \skill_i) + \lambda_{N+1} \reward_{\text{style}}(\state, \action) + \reward_{\text{reg}}(\state,  \action)
$, where $\reward_{\text{style}}$ and $\reward_{\text{reg}}$ correspond to the two types of extrinsic rewards. In practice, because the magnitudes of the individual USD and style rewards can differ, we apply exponential moving average (EMA) normalization to them to stabilize the training. Following~\cite{wang2024skildunsupervisedskilldiscovery}, we adopt value function decomposition, training a separate value function for each reward term. These are then combined using the factor weights to compute the overall advantage function, as explained in~\cref{fig:main_method}.

\textbf{Symmetry Augmentation.}
Following \citet{mittal2024symmetryconsiderationslearningtask} and ~\cref{sec:background}, we promote symmetry by augmenting the collected transitions by mirroring: \( (\state, \action, \skill,\reward) \to \{(\mirrorstate^k(\state), \mirroraction^k(\action), \mirrorskill^k(\skill), \reward)\}_{k=1}^K \).
Since the USD rewards \(\reward_{\text{USD}_i} \) are outputs of neural networks, they may not inherently respect the symmetry invariance. One way to enforce symmetry is by averaging the reward over mirrored samples: \( \frac{1}{K} \sum_{k = 1}^K \reward_{\text{USD}_i}(\mirrorstate^k(\state), \mirroraction^k(\action), \mirrorskill^k(\skill)) \). However, in practice, we found that it suffices to train all networks (actor, critics, discriminators, and encoders) on symmetry-augmented data to induce approximate symmetry in both the learned reward signal and the agent’s behavior.

The central challenge in combining USD with symmetry augmentation lies in defining \( \mirrorskill\) such that invariance in the prior distribution and composition are respected. For instance, if in the state space \(\mirrorstate^1 = \mirrorstate^2  \circ \mirrorstate^3 \) (where \(\circ\) denotes function composition), then in the skill space \(\mirrorskill^1 \) should also be equal to \( \mirrorskill^2  \circ \mirrorskill^3 \).
DIAYN priors are typically invariant to coordinate permutations, while METRA’s isotropic priors are direction-invariant. Thus, the mirroring function \(\mirrorskill \) can be realized as coordinate permutations for DIAYN and optional sign flips for METRA. More details are in~\cref{sec:app_sym}.

As a concrete example of a permutation-based skill mirroring function, consider a factored MDP with $K$ symmetries. For each state factor $i$, we assign a corresponding skill factor of dimension \( \dim(\skillspace_i) = n \cdot K\), where \(n \in \mathbb{N}^+\) is a hyperparameter. The factor skill is partitioned as \( \skill_i = [\skill_{i,1}; \cdots; \skill_{i,K}] \), with \( \dim(\skill_{i,k}) = n\).  We define \(\mirrorskill\) as permuting these \(K\)
sub-skills. A convenient choice is to let the permutations realize a Latin square \citep{denes1974latin}, where every sub-skill cycles through every position exactly once in a way that respects composition.
For a robotic agent with its four-fold symmetries, we can define the symmetry transformations as:
\begin{align}
    \mirrorskill^1(\skill_i)  &= [\skill_{i,1}; \skill_{i,2}; \skill_{i,3}; \skill_{i,4}],
    &\mirrorskill^2(\skill_i) &= [\skill_{i,3}; \skill_{i,4}; \skill_{i,1}; \skill_{i,2}], \nonumber \\
    \mirrorskill^3(\skill_i)  &= [\skill_{i,2}; \skill_{i,1}; \skill_{i,4}; \skill_{i,3}],
    &\mirrorskill^4(\skill_i) &= [\skill_{i,4}; \skill_{i,3}; \skill_{i,2}; \skill_{i,1}], \nonumber
\end{align}
which leaves the prior distribution unchanged and satisfies the desired composition rule.

Importantly, in METRA, each skill $\skill_i$ is interpreted as a direction in the learned projected state space. Shuffling coordinates or adding redundant dimensions undermines the geometric meaning and continuity of the skills. To preserve this structure, we use a low-dimensional representation (at most three coordinates \( d\leq 3 \)) per factor and define the mirroring operation to match that of the corresponding state factor. This ensures that symmetry transformations in the skill space remain consistent with those in the state space, preserving the directional semantics critical to METRA. %

\textbf{Skill Prior and Curriculum.}
For factors trained with METRA, skills are initially sampled uniformly from the unit hypersphere \(\skill_i \sim \text{U}(\mathbb{S}^{d-1}) \). Training begins with the default METRA alignment objective and gradually transitions to the norm-matching objective proposed by \citet{atanassov_constrained_2024}. The alignment objective provides a more interpretable and stable learning signal early on, facilitating initial skill acquisition. As training progresses, switching to the norm-matching objective increases the expressiveness of the skill space by allowing the skill norm to influence execution speed. During this transition, we update the skill prior by sampling variable-norm skills, enabling finer control over behavior dynamics. Additional details are in \cref{sec:app_usd}.
For DIAYN, we sample the skills from a symmetric Dirichlet prior \(\skill_i \sim \text{Dir}(\alpha)\). To emulate the separation of a categorical latent while retaining continuity, we start with a sparse prior \((\alpha_k = 0.05)\), concentrating the probability mass on a single coordinate. When the discriminator’s accuracy surpasses a preset threshold, every component is annealed linearly to 1.0, yielding a maximum-entropy Dirichlet (uniform over the \(nK-1\) dimensional probability simplex) and enabling smooth skill interpolation.

\textbf{Skill Switching.} Most USD methods fix the latent skill by sampling it once at the start of each episode. In contrast, we resample the skill multiple times within an episode. 
Without this resampling, we observe that agents tend to "lock in" to the initial skill, \ie even when the skill input is changed during deployment, the behavior remains unchanged. 
We hypothesize that this happens because the agent infers the skill from its state and, upon reaching a rewarding configuration, it chooses to stay there to maximize the USD reward. 
Resampling skills during training encourages the agent to remain responsive to the skill input, resulting in smoother skill switching at test time.

\section{Experiments}

We consider skill discovery for the quadrupedal robotic platform, ANYmal-D, which has four symmetries \citep{mittal2024symmetryconsiderationslearningtask}. We train policies on a rough terrain environment with a difficulty-based curriculum that depends on state coverage. 
The state space is factorized into base position, linear velocity, heading rate, base height, and base roll and pitch. Depending on the evaluation setup, we use different subsets of these factors and assign varying algorithm combinations to evaluate their effects.
All training is carried out in simulation using Isaac Lab~\cite{mittal2023orbit} with 2048 environments. On an NVIDIA RTX 3090, the training converges within a day.
For additional training details, please check \cref{sec:app_implementation_details}.

\begin{figure}
    \centering
    \includegraphics[width=\linewidth]{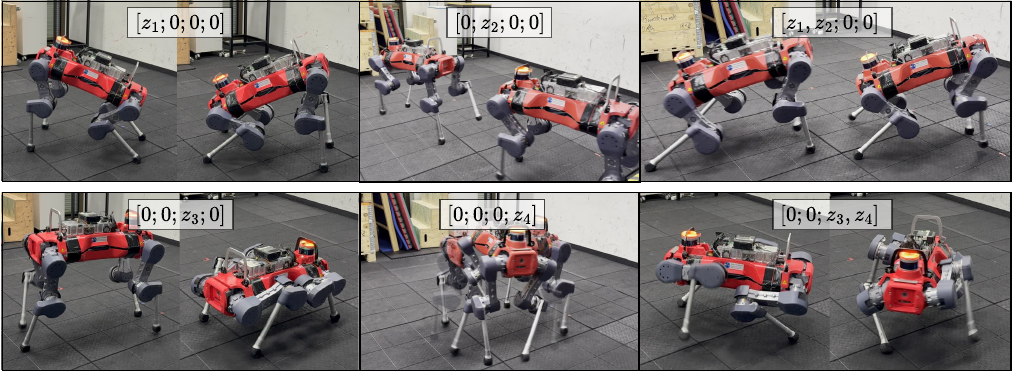}
    \vspace{-15pt}
    \caption{\textbf{Deployment of learned skills on the real robot}. The learned structured skill space enables intuitive and composable control. Each behavior corresponds to a manually commanded skill \(\skill\), set by adjusting individual skill factors \(\skill_i\). This results in diverse behaviors: pitching, walking, ducking, rotating and their combinations. Here we show walking while pitching, and ducking while rotating in the top and bottom rows respectively.}
    \vspace{-1.5em}
    \label{fig:learned_skills}
\end{figure}

\textbf{Deployment of Learned Skills.} We demonstrate the structured nature of the learned~skill space by manually commanding individual skill dimensions on the real robot. As shown in \cref{fig:learned_skills}, each skill aligns with a specific state factor and can be composed intuitively. The framework captures symmetries in behavior, for example, forward and backward walking, or tilting in opposite directions, while the style factor enables stable behaviors like standing still. Combinations of skills, such as walking while pitching or rotating while crouched, 
\begin{wrapfigure}{r}{0.5\textwidth}
  \centering
  \includegraphics[width=0.48\textwidth]{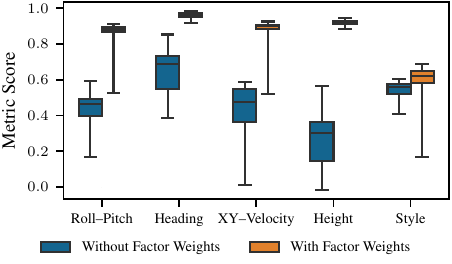}
  \vspace{-5pt}
  \caption{\textbf{Effect of factor weighting on skill metrics.} The metric score (details in \cref{app:eval_metrics}) reflects either discriminator classification accuracy (state factors) or style reward (style factor). Incorporating per-factor weights $\lambda$ enables the agent to prioritize relevant factors, yielding consistently higher scores across all dimensions.}
  \vspace{-1.5em}
  \label{fig:factor_weightingi_metrics}
\end{wrapfigure}
further highlight the composability and expressiveness of the learned skill space.

\textbf{Factor Weights.}
We evaluate the effect of per-factor weights $\lambda$ in a setup with four DIAYN-trained factors (roll-pitch, heading, planar velocity, and height) and the style factor.~\cref{fig:factor_weightingi_metrics} shows the skill discriminability and the scaled style reward. When using weights, each rollout contributes to the per-factor metrics proportionally to its assigned factor weights, normalized to avoid numerical bias. The weighted setup achieves substantially higher scores, showing that the agent learns to prioritize relevant factors. This effect is strongest with DIAYN; METRA-based variants showed smaller gains (\cref{app:exp_details}).

\looseness=-1
\textbf{Safety and Extrinsic Rewards.} To evaluate the impact of the style factor, we conduct experiments using METRA for base position and DIAYN for heading rate factor. Two sets of experiments are run, each with five different random seeds: one with the style factor active, and one with the style factor disabled by setting its weight to zero.
From \cref{tab:style_factor_eval}, we observe that the style factor significantly reduces undesirable contacts and improves discriminability for both position and heading. This suggests it promotes safer behaviors while regularizing skill learning toward more consistent, interpretable skills.

\begin{table}[t]
    \centering
    \small
    \caption{\textbf{Effect of the style factor on skill metrics and safety.} The factor metrics report classification accuracy for DIAYN and cosine similarity for METRA (with mean ± std over 5 seeds). Illegal contacts show the percentage of illegal contacts per step for different body parts.}
    \vspace{0.5em}
    \resizebox{\textwidth}{!}{
    \begin{tabular}{lccc|ccc}
        \toprule
        & \multicolumn{3}{c|}{Factor Metrics \( \uparrow \) } & \multicolumn{3}{c}{\% Illegal Contacts per Step \( \downarrow \) } \\
        \cmidrule(r){2-4}\cmidrule(r){5-7}
        Style & Extrinsic & Position & Heading & Base & Shank & Thigh \\
        \midrule
        \rowcolor[HTML]{EFEFEF} Without Style Factor & 0.19 ± 0.11 &  0.27 ± 0.14 & 0.56 ± 0.04 & 0.46 ± 0.37 & 0.75 ± 0.23 & 4.04 ± 0.06 \\
        With Style Factor                            & \textbf{0.59 ± 0.04} & \textbf{0.68 ± 0.10} & \textbf{0.72 ± 0.04} & \textbf{0.00 ± 0.00} & \textbf{0.12 ± 0.07} & \textbf{0.03 ± 0.00} \\
        \bottomrule
    \end{tabular}
    }
    \label{tab:style_factor_eval}
    \vspace{-0.5em}
\end{table}

\textbf{Comparing Different USD Objectives.} To evaluate the flexibility and effect of assigning different USD algorithms to individual state factors, we compare diversity across various algorithm configurations. We factorize the state into base position and heading rate.
In our setup, we use METRA for position, favoring broad state-space coverage, and DIAYN for heading, encouraging skill separability.  We compare this mixed setup against several baselines: both factors trained with DIAYN (similar to DUSDi \citep{hu_disentangled_2024} but with a style factor and regularization, denoted as ``DUSDi" in the tables), both with METRA (``2×METRA"), and single-objective baselines where the two factors are combined into one (i.e., no factorization) and trained with either DIAYN~\citep{eysenbach_diversity_2018} or METRA~\citep{park_metra_2024}, both with style and regularization.
For evaluating diversity, we follow \citet{zahavy2023discoveringpoliciesdominodiversity_domino} and compute Monte Carlo estimates of successor representations for each skill and report their standard deviation in \cref{tab:alg_mixing_results}.
Higher values indicate broader diversity in the corresponding factor. 
We observe that METRA significantly improves diversity in the position factor, while DIAYN achieves higher diversity for the heading factor. The mix of algorithms per factor outperforms using a single algorithm for all factors.
\begin{table}[t]
    \centering
    \small
    \caption{\textbf{Comparison against different USD approaches across state factors.} Diversity is measured as state coverage (details in \cref{app:eval_metrics}) with mean \(\pm\) std over 5 seeds. Higher values indicate broader skill coverage.}
    \vspace{0.5em}
    \begin{tabular}{lcc|cc}
        \toprule
        & \multicolumn{2}{c|}{Algorithm Chosen per Factor } & \multicolumn{2}{c}{Diversity per Factor \( \uparrow \)} \\
        \cmidrule(r){2-3} \cmidrule(r){4-5}
        Approach & Position & Heading & Position & Heading\\
        \midrule
        \rowcolor[HTML]{EFEFEF} DIAYN & \multicolumn{2}{c|}{DIAYN (\(\text{dim}(\skill)=8\))}  & \textcolor{black!30!}{0.389 ± 0.183} & \textcolor{black!67!}{1.067 ± 0.532} \\
        METRA & \multicolumn{2}{c|}{METRA (\(\text{dim}(\skill)=3\))} & \textcolor{black!100!}{9.832 ± 0.808} & \textcolor{black!30!}{0.212 ± 0.018} \\
        \rowcolor[HTML]{EFEFEF} DUSDi & DIAYN (\(\text{dim}(\skill_1)=4\)) & DIAYN (\(\text{dim}(\skill_2)=2\)) & \textcolor{black!37!}{1.363 ± 0.333} & \textcolor{black!100!}{1.811 ± 0.084} \\
        2xMETRA & METRA (\(\text{dim}(\skill_1)=2\)) & METRA (\(\text{dim}(\skill_2)=1\)) & \textcolor{black!93!}{8.836 ± 1.411} & \textcolor{black!33!}{0.271 ± 0.078} \\ \midrule
        \rowcolor[HTML]{EFEFEF} Mixed (Ours) & METRA (\(\text{dim}(\skill_1)=2\)) & DIAYN (\(\text{dim}(\skill_2)=2\)) & \textcolor{black!92!}{8.776 ± 0.667} & \textcolor{black!66!}{1.031 ± 0.476} \\
        \bottomrule
    \end{tabular}

    \label{tab:alg_mixing_results}
    \vspace{-1em}
\end{table}

\begin{figure}[t]
  \centering
  \subfloat{\includegraphics[width=.48\columnwidth]{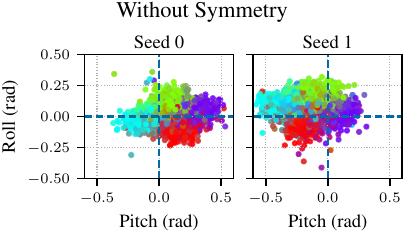}}
  \hfill
  \subfloat{\includegraphics[width=.48\columnwidth]{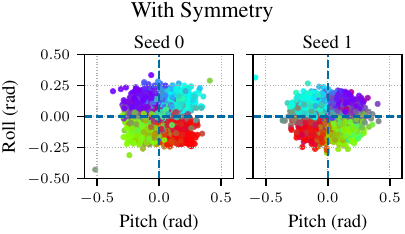}}
  \caption{\textbf{Impact of symmetry augmentation on skill-to-state mappings.} Each point shows roll and pitch angles (in radians) reached by the policy, colored by the commanded skill. Without symmetry augmentation, the mapping is arbitrary and less structured. With symmetry augmentation, skills align symmetrically across the factor space, leading to more interpretable and balanced behaviors.}
  \label{fig:sym_space_skills}
  \vspace{-1em}
\end{figure}

\textbf{Symmetric Skill Discovery.} To evaluate the effect of symmetry augmentation, we train policies with and without symmetry bias. We observe that symmetry augmentation does not result in faster convergence or higher evaluation metrics. Additionally, policies trained with symmetry augmentation often obtain lower metric scores for factors trained with METRA-style rewards and perform similarly with DIAYN-style rewards (see \cref{sec:app_additional_experiments}). This could be due to METRA-based factors being harder to symmetrize effectively due to the geometric interpretation of the skill, or due to the augmentation technique being suboptimal compared to more structured methods like Latin Square symmetry.
Nevertheless, symmetry augmentation leads to more interpretable and structured skill-to-state mappings. For example, when controlling base heading, policies with symmetry learn to rotate uniformly in both directions, whereas for those without symmetry, the behavior often tends to be biased. In \cref{fig:sym_space_skills}, we visualize the effect of symmetry augmentation on state space coverage, showing how learned skills become more symmetrically distributed.

\textbf{Downstream Task.} We assess the utility of the learned skill libraries on a rough-terrain waypoint–navigation task with random position goals (up to 15\,$\text{m}$ away) and random target headings. We compare three control schemes: \textbf{(i)} \textit{Direct}, a basic PPO policy that outputs joint position commands, \textbf{(ii)} \textit{Oracle}, a hierarchical PPO policy with a hand-tuned velocity-tracking controller~\citep{pmlr-v164-rudin22a} as the low-level policy, and \textbf{(iii)} Skill-based, the same hierarchy as the oracle but using pre-trained skill-conditioned policies (from \cref{tab:alg_mixing_results}) as the low-level controller.
As shown in \cref{tab:downstream_metrics}, our \textit{mixed} setup (using METRA for position, DIAYN for heading) closely matches the oracle, achieving low tracking error and high success rate. In contrast, direct control fails entirely because of poor structure in the action space. Mismatched or single-objective USD skill libraries also underperform, underscoring the importance of appropriate factor–algorithm combinations for downstream performance. Additional implementation details are in \cref{sec:app_implementation_details}.

\begin{table}[t]
    \centering
    \footnotesize
    \caption{\textbf{Performance on downstream navigation task.}  Metrics include average reward, tracking errors, and episode termination ratios: timeouts (exceeding 30s), base collisions, or successful goal-reaching.}
    \vspace{0.5em}
    \resizebox{\textwidth}{!}{
    \begin{tabular}{l|c|cc|ccc}
        \toprule
        Approach & Reward \( \uparrow \) & \multicolumn{2}{c|}{Tracking Error} & \multicolumn{3}{c}{Termination Ratio} \\
         \cmidrule(r){3-4} \cmidrule(r){5-7}
        &  & Heading \( \downarrow \) & Position \( \downarrow \)  & Goal Reached \( \uparrow \) & Base Collision \( \downarrow \) & Time Out  \\
        \midrule
        \rowcolor[HTML]{EFEFEF} Direct         & \textcolor{black!30!}{1.85 ± 0.48} & \textcolor{black!30!}{1.56 ± 0.03} & \textcolor{black!30!}{11.00 ± 0.12} & \textcolor{black!30!}{0.004 ± 0.000} & \textcolor{black!30!}{0.996 ± 0.106} & \textcolor{black!100!}{0.000 ± 0.000} \\
        DIAYN                                  & \textcolor{black!41!}{27.87 ± 2.42} & \textcolor{black!59!}{1.34 ± 0.05} &  \textcolor{black!55!}{7.50 ± 0.26} & \textcolor{black!32!}{0.034 ± 0.000} & \textcolor{black!100!}{0.000 ± 0.000} & \textcolor{black!30!}{0.966 ± 0.000} \\
        \rowcolor[HTML]{EFEFEF} DUSDi          & \textcolor{black!44!}{35.26 ± 3.65} & \textcolor{black!56!}{1.36 ± 0.05} &  \textcolor{black!61!}{6.74 ± 0.31} & \textcolor{black!33!}{0.039 ± 0.005} & \textcolor{black!100!}{0.001 ± 0.001} & \textcolor{black!31!}{0.960 ± 0.004} \\
        2xMETRA                                & \textcolor{black!43!}{32.05 ± 7.73} & \textcolor{black!33!}{1.54 ± 0.04} &  \textcolor{black!58!}{7.05 ± 0.32} & \textcolor{black!37!}{0.090 ± 0.052} & \textcolor{black!58!}{0.593 ± 0.103} & \textcolor{black!77!}{0.317 ± 0.032} \\
        \rowcolor[HTML]{EFEFEF} METRA          & \textcolor{black!64!}{81.62 ± 50.20} & \textcolor{black!60!}{1.33 ± 0.14} &  \textcolor{black!76!}{4.68 ± 2.78} & \textcolor{black!54!}{0.300 ± 0.234} & \textcolor{black!73!}{0.378 ± 0.532} & \textcolor{black!76!}{0.322 ± 0.037} \\
        Mixed (Ours)                           & \textcolor{black!93!}{148.55 ± 29.24} & \textcolor{black!100!}{1.03 ± 0.14} & \textcolor{black!100!}{1.33 ± 0.27} & \textcolor{black!94!}{0.797 ± 0.424} & \textcolor{black!99!}{0.012 ± 0.014} & \textcolor{black!86!}{0.191 ± 0.185} \\
        \rowcolor[HTML]{EFEFEF} Oracle         & \textcolor{black!100!}{164.37 ± 21.42} & \textcolor{black!95!}{1.07 ± 0.17} & \textcolor{black!97!}{1.66 ± 0.65} & \textcolor{black!100!}{0.871 ± 0.427} & \textcolor{black!96!}{0.052 ± 0.031} & \textcolor{black!94!}{0.078 ± 0.111} \\
        \bottomrule
    \end{tabular}
    }
    \vspace{-1em}
    \label{tab:downstream_metrics}
\end{table}

\section{Conclusion}
\label{sec:conclusion}

\looseness=-1
We presented a modular framework for unsupervised skill discovery (USD) that employs user-defined factorization of the state space and allows assigning different algorithms to each factor. This design leverages the complementary strengths of USD objectives: METRA excels at exploring unbounded dimensions like position through latent-space traversal, while DIAYN produces more distinguishable behaviors on bounded dimensions like heading or orientation.
To support real-world deployment, our framework introduces several key components: a style factor and regularization terms that encourage safe and stable behaviors; symmetry augmentation that induces morphology-aware structure; and a factor-weighting mechanism that prioritizes relevant behaviors and resolves conflicts across active skills. We showed that these components individually contribute to skill quality, and their combination enables a smooth zero-shot transfer from simulation to hardware as well as intuitive control through direct skill commands.
On downstream navigation tasks, our approach achieves near-oracle performance and significantly outperforms single or mismatched USD setups, demonstrating improved sample efficiency. The framework integrates seamlessly with scalable simulation and on-policy RL, and remains compatible with any USD method with intrinsic rewards. We open-source the code for future research in this direction: \href{https://leggedrobotics.github.io/d3-skill-discovery}{https://leggedrobotics.github.io/d3-skill-discovery/}.

\looseness=-1
Future directions include scaling to more complex behaviors such as loco-manipulation and climbing boxes, which may require stronger exploration or curricula. Additionally, extending the framework to different robots with varying morphologies and symmetries is another promising avenue.

\section{Limitations}

Our results show that factorized, symmetry-aware USD can produce safe, deployable skills, but several gaps remain.

\begin{figure}[!ht]
    \centering
    \begin{subfigure}[b]{0.48\textwidth}
        \centering
        \includegraphics[width=0.9\linewidth]{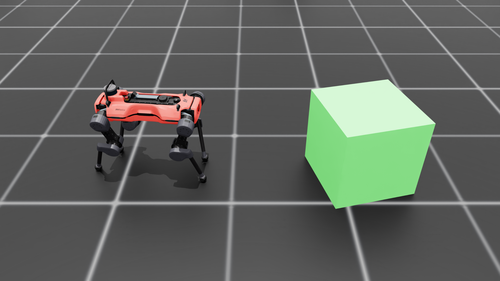}
        \caption{Flat terrain with a \SI{0.5}{\meter} cubic box weighing \SI{10}{\kilogram}.
        }
        \label{fig:terrain_box}
    \end{subfigure}
    \hfill
    \begin{subfigure}[b]{0.48\textwidth}
        \centering
        \includegraphics[width=0.9\linewidth]{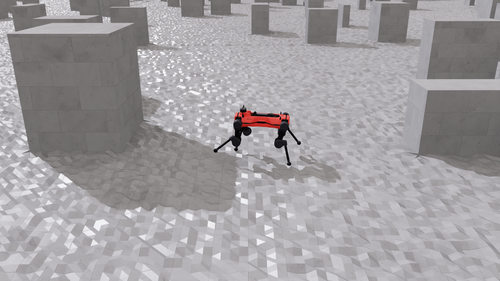}
        \caption{Terrain with randomly placed static obstacles.}
        \label{fig:terrain_obstacle_curr}
    \end{subfigure}
    \hfill
    \caption{Environments for more complex skill discovery. (a) Adding a box pose factor encourages pushing, but the agent relies on unsafe, forceful collisions rather than controlled manipulation. (b) In an obstacle-rich environment, the agent explores but fails to discover safe avoidance behaviors without explicit rewards.}
    \vspace{-0.35em}
    \label{fig:additional_experiment_terrains}
\end{figure}

\textbf{Discovering More Complex Skills.} While the framework handled rough-terrain locomotion, extending it to complex, interaction-rich tasks was considerably more challenging. For instance, when adding a ``box pose'' factor to encourage loco-manipulation (\cref{fig:terrain_box}), the agent seldom learned the multi-stage behavior of walking to the box before the interaction. The discovered pushing skills often relied on unsafe, forceful collisions. Similarly, for obstacle-rich navigation without task-rewards (\cref{fig:terrain_obstacle_curr}), the agent failed to acquire obstacle avoidance behaviors. These observations indicate that additional guidance, such as task-aware curricula or alternative intrinsic objectives, is required to unlock more complex loco-manipulation and locomotion skills. More details are provided in~\cref{sec:app_additional_experiments}.

\textbf{Cost of Symmetry in Skill Emergence.}
To enable the robot to learn pedipulation-like behaviors~\cite{arm2024pedipulate}, we added factors for the position of each foot of the robot. However, we observed that the symmetry-mirroring suppressed the emergence of lifting of individual feet. Disabling symmetry produced a lift for only one of the legs, but this behavior did not extend to the other legs of the robot. These results suggest that fine-scale skills may benefit from softer symmetry biases or an explicit per-leg curriculum for skill discovery.

\textbf{User-defined Design Decisions.} Our proposed method still requires user decisions about factorization, algorithm selection, hyperparameters, and safety shaping. Automating these design choices and adding hard safety guarantees would make the framework more plug-and-play, increasing its usability across diverse robotic platforms and downstream tasks.

\acknowledgments{
We thank Philip Arm and David Hoeller for helpful discussions at the start of the project.
This research was supported by the Swiss National Science Foundation through the National Centre of Competence in Automation (NCCR automation).
Additionally, it has received funding from the European Union’s Horizon Europe Framework Programme under grant agreement No 101121321. 
}

\bibliography{bibliography.bib}  %

\cleardoublepage

\appendix
\section{Appendix}

\subsection{Unsupervised Skill Discovery Algorithms}
\label{sec:app_usd}

\paragraph{DIAYN: Diversity Is All You Need.}
DIAYN~\citep{eysenbach_diversity_2018} aims to learn a skill-conditioned policy \( \policy(\action \mid \state, \skill) \) by maximizing the mutual information (MI) between states and skills \(  I(\statevar; \skillvar) \triangleq \KL{p(\state, \skill)}{p(\state) p(\skill)} \equiv \entropy(\skillvar) - \entropy(\skillvar \mid \statevar) \), where \( \entropy(\cdot) \) denotes the Shannon or differential entropy. Intuitively, minimizing \( \entropy(\skillvar \mid \statevar) \) means that the skill $\skill$ should be easy to infer given the state $\state$.
This is implemented by learning a discriminator \( \discriminator_{\phi}(\state \mid \skill) \) that approximates the posterior \(p(\state \mid \skill) \).
The policy and discriminator form a cooperative game: the discriminator predicts the skill that led to the policy visiting certain states, while the policy seeks to visit states that make it easy for the discriminator to identify the skill.
The resulting reward is:
\begin{align}
    \reward_{\text{DIAYN}}(\state,  \action, \skill) = \log \discriminator_{\phi}(\skill \mid \state) - \log p(\skill)
\end{align}

Note, \( \log p(\skill) \) only needs to be included in the reward term if \( p(\skill) \) is not uniform.
The discriminator is trained by maximizing the log-likelihood of the posterior $
   \mathbb{E}_{\skill \sim p(\skill), \state \sim \policy(\skill)} \log \discriminator_{\phi}(\skill \mid \state).
$
In practice, the discriminator predicts parameters of a distribution over the skill space, which depends on the selection of the prior \( \prior(\skill) \).

The authors of the original paper use a categorical distribution for the prior \( \prior(\skill) \). and noted that learning with continuous skill distributions such as uniform or Gaussian distribution degrades performance.
\begin{wraptable}{r}{0.5\textwidth}
    \centering
    \caption{\textbf{Possible choices for skill distributions.} Depending on the choice of the prior distribution for the skills, we choose the posterior according to this table.}
    \label{tab:distr_types}
    \begin{tabular}{l l}
        \toprule
        Prior \(\prior(\skill)\) & Posterior \(\discriminator(\skill \mid \state)\) \\ \midrule
        Uniform categorical &  Categorical \\
        Uniform continuous & Gaussian \\
        Gaussian \( \mathcal{N}(\mathbf{0}, \mathbf{I}) \)& Gaussian \\
        Uniform on sphere & Von Mises-Fisher \\
        Symmetric Dirichlet & Dirichlet \\ \bottomrule
    \end{tabular}
\end{wraptable}
\citet{imagawa2023unsuperviseddiscoverycontinuousskills} show that to learn more skills, using a continuous distribution yields better results than using a large number of discrete skills. Depending on the selection of the skill prior, the parameterization of the discriminator needs adjustments. Importantly, the support of the posterior has to contain the support of the prior. In \cref{tab:distr_types} we list different combinations of priors and posteriors we tested. We found that Dirichlet-distributed skills offer a good trade-off between continuous skill expressiveness and discriminability accuracy.

\paragraph{METRA: Metric-Aware Abstraction.}
METRA~\citep{park_metra_2024} (as well as LSD~\citep{park2022lipschitzconstrainedunsupervisedskilldiscovery} and CSD~\cite{atanassov_constrained_2024}) aims to learn a skill-conditioned policy by learning an encoder \( \staterepr \) that maps states into a latent space of the same dimensionality as the skill space. The skill discovery objective is to maximize the alignment of latent transitions \( \staterepr(\state') - \staterepr(\state) \) with the skill \( z \) under a constraint  \( \| \staterepr(\state) - \staterepr(\state') \|_2 \leq d(\state', \state) \), with distance metric \( d(\cdot, \cdot) \). LSD proposes to use the Euclidean distance between the states, \(  d(\state', \state) = \| \state' - \state \| \); CSD \citep{park2023controllabilityawareunsupervisedskilldiscovery} proposes to use controllability-aware distance metric; while METRA~\citep{park_metra_2024} proposes to use temporal distance, \ie the minimum number of episodic steps to reach $\state'$ from $\state$, which in their setup is simply \( d(\state', \state) = 1\).

The objective of the encoder in METRA is defined as, for all $ (\state, \state') \in \statespace_{\text{adj}}$:
\begin{align}
\mathcal{J}_\text{METRA}(\theta, \staterepr) = \mathbb{E}_{\skill \sim p(\skill), \state \sim \policy(\skill)} \left[ (\staterepr(\state') - \staterepr(\state))^\top \skill \right] \text{s.t.} \| \staterepr(\state) - \staterepr(\state') \|_2 \leq d(\state', \state).
\end{align}
In practice, this constrained obejctive is optimized via dual-gradient descent. It simplifies into a reward function that rewards the agent if its actions result in state transitions that align with the skill in the latent state space:
\begin{align}
    \reward_{\text{METRA}}(\state, \skill, \state') = (\staterepr(\state') - \staterepr(\state))^\top \skill.
\end{align}

However, this objective leads to skills that move maximally fast through the state space, which might not be desirable. Instead, \citet{atanassov_constrained_2024} propose a norm-matching objective to also control the execution speed of a skill, where for all $ (\state, \state') \in \statespace_{\text{adj}}$:
\begin{align}
\mathcal{J}_\text{NM}(\theta, \staterepr) =\mathbb{E}_{\skill \sim p(\skill), \state \sim \policy(\skill)} \left[ \|(\staterepr(\state') - \staterepr(\state) - \skill \|^2 \right] \text{s.t.} \| \staterepr(\state) - \staterepr(\state') \| \leq d(\state', \state).
\end{align}
The resulting reward becomes a function of the error between the skill and the latent transition:
\begin{align}
    \reward_{\text{METRA,nm}}(\state, \skill, \state') = (1 + \sigma \|(\staterepr(\state_{t+1}) - \staterepr(\state_{t})) - \skill \|_2^2)^{-1}
\end{align}
where $\sigma \in \mathbb{R}$ is a scaling factor.

In practice, we found that the norm-matching objective has a considerably weaker alignment component, making it difficult to train from scratch, particularly when the initial alignment between latent state transitions and skills is poor. The original alignment objective is often more stable in the early stages of training. To combine the strengths of both objectives, we implement a curriculum that starts with the original alignment objective and smoothly transitions to the norm-matching objective as alignment performance improves, \ie the final objective is a weighted sum:
\begin{align}
    \mathcal{J}_\text{METRA mix}(\theta, \staterepr) = (1-\alpha_\text{mix}) \mathcal{J}_\text{METRA}(\theta, \staterepr) + \alpha_\text{mix} \mathcal{J}_\text{NM}(\theta, \staterepr).
\end{align}
The transition is controlled by the interpolation parameter \(\alpha_\text{mix} \in [0, 1]\), which is dynamically calculated based on the cosine similarity between the latent state transition \((\staterepr(\state') - \staterepr(\state))\) and the skill \(\skill\). Specifically, \(\alpha_\text{mix}\) linearly ramps from 0 to 1 as the cosine similarity score increases over a predefined range. For our experiments, this objective switching range is set to \([0.5, 0.7]\), as detailed in \cref{tab:Metra_params}. This means that when the cosine similarity is below 0.5, the agent is trained purely on the alignment objective (\(\alpha_\text{mix}=0\)), and when it exceeds 0.7, the training switches completely to the norm-matching objective (\(\alpha_\text{mix}=1\)). In between these values, the objectives are mixed linearly.
The weight \(\alpha_\text{mix}\) is used for both the agent's reward calculation and the encoder's loss function.

\paragraph{DUSDi: Disentangled Unsupervised Skill Discovery.}
To learn disentangled skills, \citet{hu_disentangled_2024} propose learning two discriminators per factor, based on DIAYN. The first discriminator predicts the skill factor from the respective state factor \( \discriminator_i(\skill_i \mid \state_i) \), while the second discriminator predicts the skill from every other state factor: \( \discriminator_{\neg i}(\skill_i \mid \state_{\neg i}) \), where \( \state_{\neg i} \in \statespace_ {\neg i} = \statespace_1 \times \dots \statespace_{i-1} \times \statespace_{i+1} \times \dots \statespace_N \). This results in a reward function defined as:
\begin{equation}
\reward_{\text{DUSDI}}(\state,  \action, \skill) \triangleq \sum_{i=1}^N \discriminator_i(\skill_i \mid \state_i) - \gamma  \discriminator_{\neg i}(\skill_i \mid \state_{\neg i}),
\end{equation}
where \( \gamma < 1 \) is a hyperparameter that controls the importance of the entanglement penalty relative to the skill-factor association (typically $\gamma $ = 0.1). The first reward component is the standard DIAYN reward, while the second one has the same formulation but is used as a penalty. The harder it is to infer a skill factor given other state factors, the more disentangled the learned skills are.

\subsection{Symmetry Augmentation}
\label{sec:app_sym}

Symmetry augmentation can help boost sample efficiency and learn smoother behaviors. \citet{mittal2024symmetryconsiderationslearningtask} propose to simply augment the collected data instead of introducing an extra symmetry objective, or enforcing symmetry in the network architecture.

So far, symmetry biases have not been used as part of unsupervised skill discovery. However, it might be useful to learn symmetric skills and boost exploration. To utilize symmetry augmentation, the MDP needs to have symmetries, which requires the reward to be invariant to symmetry transformations. In general, this is not the case in skill discovery. One way to enforce symmetry in the reward is computing it as an average over all symmetries:
\begin{align}
    \reward_{\text{sym}}(\state, \action, \skill) = \frac{1}{K} \sum_{i = 1}^K \reward(\mirrorstate^i(\state), \mirroraction^i(\action), \mirrorskill^i(\skill)), 
\end{align}
where \(\reward_{\text{sym}}\) is a reward that is guaranteed to be invariant over all symmetries.
However, in practice, we found that it suffices to train all networks on symmetry-augmented data.

\paragraph{Skill Mirroring Function Properties.}
The function that mirrors skills, \( \mirrorskill \), can be chosen freely as long as it preserves: (i) the invariance of the skill prior and (ii) the group composition of the underlying symmetries.
The choice of a valid mirroring function \( \mirrorskill \) depends on the skill prior \(p(\skill)\). For METRA, we use an isotropic prior, where all skills with the same norm \(\|\skill\|_2\) have the same probability, regardless of their direction. This means any mirroring function \( \mirrorskill \) is valid as long as it is norm-preserving (e.g., a reflection or rotation). For DIAYN, we use a symmetric Dirichlet distribution, \(\text{Dir}(\skill \mid \alpha)\), where all entries of the concentration parameter \(\alpha \in \mathbb{R}_+^d\) are equal. The resulting probability distribution is invariant to any permutation of the entries of \(\skill\). Therefore, for DIAYN, any mirroring function \( \mirrorskill \) is valid as long as it only permutes the entries of the skill vector.

The set of mirroring functions must also respect the composition of the physical symmetries. For example, mirroring a state \emph{left--right} followed by \emph{front--back} should yield the same physical transformation as a \(180^{\circ}\) rotation about the \(z\)-axis. The same composition must hold in the skill space. Otherwise, symmetry augmentation introduces contradictory training signals. Concretely, if
$
\mirrorstate^{1}\!\bigl(\mirrorstate^{2}(\state_i)\bigr)=\mirrorstate^{3}(\state_i)
\quad\forall\,\state_i\in\statespace_i,
$
but there exists a skill
$
\skill_i\in\skillspace_i\quad\text{s.t.}\quad
\mirrorskill^{1}\!\bigl(\mirrorskill^{2}(\skill_i)\bigr)\neq\mirrorskill^{3}(\skill_i),
$
then symmetry augmentation can produce tuples with the \emph{same} state but \emph{different} mirrored skills:
\[
\bigl(\mirrorstate^{3}(\state_i),\;\dots,\;\mirrorskill^{1}\!\bigl(\mirrorskill^{2}(\skill_i)\bigr)\bigr)
\;\;\text{and}\;\;
\bigl(\mirrorstate^{3}(\state_i),\;\dots,\;\mirrorskill^{3}(\skill_i)\bigr).
\]
Since states are now paired with ambiguous skills, any skill-conditioned reward or discriminator cannot remain consistent, yielding irreconcilable gradients and hindering learning. Therefore, \( \mirrorskill \) must satisfy \( \mirrorskill^{1}\!\circ\mirrorskill^{2} = \mirrorskill^{3} \), to ensure coherent symmetry-augmented training.

The quadruped ANYmal-D is left-right and front-back symmetric, resulting in four symmetry transformations: identity, left-right reflection, front-back reflection, and their composition, a \(180^\circ\) rotation about the z-axis.

\paragraph{Skill Mirroring Function Implementation.}
For \textbf{DIAYN}, we mirror skills such that subskills form a Latin square.
We use the following skill permutations:
\begin{align*}
    \mirrorskill^1(\skill_i) &=  [\skill_i^1; \skill_i^2; \skill_i^3; \skill_i^4] \\
    \mirrorskill^2(\skill_i) &=  [\skill_i^3; \skill_i^4; \skill_i^1; \skill_i^2] \\
    \mirrorskill^3(\skill_i) &=  [\skill_i^2; \skill_i^1; \skill_i^4; \skill_i^3] \\
    \mirrorskill^4(\skill_i) &=  [\skill_i^4; \skill_i^3; \skill_i^2; \skill_i^1]
\end{align*}

Permuting sub-skills gives the latent space room for states that are invariant to certain symmetries. Let \(\statespace_{\text{sym} (i,j)} = \bigl\{ \state \in \statespace \mid \; \mirrorstate^i(\state) = \mirrorstate^j(\state) \bigr\} \) be such states (e.g., forward/backward velocity is invariant to a left–right flip). Whenever \( \|\statespace_{\text{sym} (i,j)} \|>1 \) is, we require matching skills \( \skillspace_{\text{sym} (i,j)} = \bigl\{ \skill \in \skillspace \mid \; \mirrorskill^i(\skill) = \mirrorskill^j(\skill) \bigr\} \),
which our permutation-based mirroring provides automatically:
\begin{align*}
\begin{array}{c|cccc}
\skillspace_{\text{sym}(i,j)} & j=1 & j=2 & j=3 & j=4 \\
\hline
i=1 & \skillspace & [a; b; a; b] & [a; a; b; b] & [a; b; b; a] \\
i=2 & [a; b; a; b] & \skillspace & [a; b; b; a] & [a; a; b; b] \\
i=3 & [a; a; b; b] & [a; b; b; a] & \skillspace & [a; b; a; b] \\
i=4 & [b; a; a; b] & [a; a; b; b] & [a; b; a; b] & \skillspace \\
\end{array}
\end{align*}
where \(a, b \in \mathbb{R}^n\) denote subskills.
This pattern guarantees that a state symmetric under \(i\) and \(j\) can always be paired with a unique skill lying in \(\skillspace_{\text{sym}(i,j)} \). 

If \( \statespace_{\text{sym} (i,j)} = \statespace \), i.e., every state of a factor is invariant under a pair \((i,j)\), then the skill map must satisfy \( \skillspace_{\text{sym} (i,j)} = \skillspace \) as well.
For the heading-rate factor, a scalar that flips sign under left-right or front-back reflections but is unchanged by their composition, both reflections are merged into a single ``flip", resulting in the symmetries \( \{1,2\} =\) (identity, flip). With two subskills we set \(\mirrorskill^1(\skill_i) =  [\skill_i^1; \skill_i^2]\) and \(\mirrorskill^2(\skill_i) = [\skill_i^2; \skill_i^1]\). For states with no symmetries, e.g., the base height, we omit symmetry augmentation for the skills completely.

For \textbf{METRA}, we treat the skill vector \(\skill_i\) as a directional proxy for its associated state factor and mirror it accordingly. This approach is grounded in the geometric nature of METRA's alignment objective.
Specifically, since the skill \(\skill_i\) represents a direction, the mirroring function \(\mirrorskill^k\) applied to the skill is defined to be the same geometric transformation as the function \(\mirrorstate^k\) applied to the state. For instance, if a state factor like the robot's base velocity is reflected across a plane for a left-right symmetry transformation, its corresponding skill vector is also reflected across that same plane.
Applying identical symmetry transformations to both states and skills introduces a strong inductive bias that aligns the learned skill behaviors with the physical symmetries of the robot. This consistency is critical for the METRA objective, which directly rewards the alignment between the skill vector and the change in the latent state representation.

In our experiments, we also found it crucial to limit the dimensionality of these directional skill vectors to \(d \le 3\). This is an empirical finding. When we experimented with higher-dimensional skill vectors (\(d > 3\)), the policy consistently learned to ignore the additional, non-mirrored dimensions. This behavior effectively caused the learned skill space to collapse back into a 3D geometric subspace, and attempts to define more complex, higher-dimensional mirroring functions did not prevent this instability. Therefore, constraining the skill dimensionality to match the 3D nature of the physical transformations proved to be the most stable and effective approach.

\subsection{Implementation Details}
\label{sec:app_implementation_details}

\paragraph{Policy Network.}  For the policy, we use a 3-layer MLP [512, 256, 128] with elu activations. The action space is 12-dimensional, corresponding to the robot's joint position targets. For each action dimension, the policy predicts the mean and log std of a Gaussian distribution, of which we clamp the standard deviation to the range \([e^{-5}, e^2] \). During training, actions are sampled from the predicted distribution. During deployment, the action is the predicted mean. %

\paragraph{Hyperparameters.}  We list hyperparameters for PPO in \cref{tab:PPO_params}, for METRA in \cref{tab:Metra_params}, and for DIAYN in \cref{tab:Diayn_params}. The rewards for the style factor are listed in \cref{tab:style_rewards} and the regularization penalties in \cref{tab:regularization_rewards}. The policy observations can be found in \cref{tab:policy_observations}.

\begin{table}[htb]
    \centering
    \begin{minipage}[t]{0.48\textwidth}
        \centering
        \caption{PPO Hyperparameters}
        \begin{tabular}{ll}
            \toprule
            Hyperparameter & Value \\
            \midrule
            PPO clip ratio & 0.2   \\
            Value clip ratio & 0.2   \\
            Num env steps before update & 24   \\
            Num learning epochs & 5 \\
            Num minibatches & 4 \\
            Learning rate & 1.0e-3 \\
            Discount factor & 0.99 \\
            GAE lambda & 0.95   \\
            KL target & 0.01   \\
            Max grad norm & 1.0   \\
            \bottomrule
        \end{tabular}
        \label{tab:PPO_params}
    \end{minipage}%
    \hfill
    \begin{minipage}[t]{0.48\textwidth}
        \centering
        \caption{Policy Observations}
        \begin{tabular}{ll}
            \toprule
            Name & Dim \\
            \midrule
            Base xy-position in world frame & 2    \\
            Base linear velocity & 3    \\
            Base angular velocity & 3    \\
            Projected gravity & 3    \\
            Previous action & 12    \\
            Joint position & 12    \\
            Joint velocity & 12    \\
            Height scan (\SI{1.6}{\meter}$\times$\SI{1.0}{\meter}) & 231    \\
            \bottomrule
        \end{tabular}
        \label{tab:policy_observations}
    \end{minipage}
\end{table}

\begin{table}[htb]
    \centering
    \begin{minipage}[t]{0.48\textwidth}
        \centering
        \caption{METRA Hyperparameters}
        \begin{tabular}{ll}
            \toprule
            Hyperparameter & Value \\
            \midrule
            Learning rate & 1.0e-4   \\
            Initial Lagrange multiplier & 30.0   \\
            Lagrange multiplier lr & 1e-4   \\
            Lagrange multiplier slack & 1e-5   \\
            Objective switching range & (0.5, 0.7) \\
            Network & MLP: [256, 256] \\
            Norm-matching \(\sigma\) & 10.0 \\
            \bottomrule
        \end{tabular}
        \label{tab:Metra_params}
    \end{minipage}%
    \hfill
    \begin{minipage}[t]{0.48\textwidth}
        \centering
        \caption{DIAYN Hyperparameters}
        \begin{tabular}{ll}
            \toprule
            Hyperparameter & Value \\
            \midrule
            Skill distribution & Dirichlet   \\
            Disentanglement \(\lambda\) & 0.1   \\
            Network & MLP: [256, 256] \\
            Learning rate & 1e-4   \\
            Dirichlet param range & (0.05, 1.0) \\
            \bottomrule
        \end{tabular}
        \label{tab:Diayn_params}
    \end{minipage}
\end{table}

\begin{table}[htb]
    \centering
    \caption{Style Factor Rewards}
    \begin{tabular}{lll}
        \toprule
        Name & Objective & Weight \\
        \midrule
        Joint torques & \( \|\boldsymbol{\tau}  \|_2^2 \) & -1.0e-3  \\
        Joint acceleration & \( \|\ddot{\mathbf{q}} \|_2^2 \) & -1.0e-5  \\
        Action rate & \( \| \action_t - \action_{t-1} \|_2^2 \) & -0.2  \\
        Action norm & \( \| \action \|_2^2 \) & -0.4  \\
        Undesired Contacts & \(\sum_{b \in \{ \text{Thighs},\ \text{Shanks},\ \text{Base} \}} \mathbf{1} \left[ \text{contact}(b) \right]\) & -30.0 \\
        Base height & \( \| p_z - 0.55 \|^2 \) & -10.0  \\
        Flat orientation & \( \left\| \mathbf{g}_{b,\text{xy}} \right\|_2^2\) & -10.0  \\
        \bottomrule
    \end{tabular}
    \vspace{0.5em}
    \label{tab:style_rewards}
\end{table}
\begin{table}[ht]
    \centering
    \caption{Regularization Rewards}
    \begin{tabular}{lll}
        \toprule
        Name & Objective & Weight \\
        \midrule
        Joint torques & \( \|\boldsymbol{\tau}  \|_2^2 \) & -1.0e-3  \\
        Joint acceleration & \( \|\ddot{\mathbf{q}} \|_2^2 \) & -2.5e-7  \\
        Action rate & \( \| \action_t - \action_{t-1} \|_2^2 \) & -0..05  \\
        Torque limits & 
        \( \| \max\left( \boldsymbol{\tau} - \tau_{\max},\ 0 \right) \|_1+\| \min\left( \boldsymbol{\tau} - \tau_{\min},\ 0 \right) \|_1
        \) & -15.0 \\
        Torque ratio limits & 
        \( \| \max\left( \boldsymbol{\tau} - 0.75 \tau_{\max},\ 0 \right) \|_1+\| \min\left( \boldsymbol{\tau} - 0.75 \tau_{\min},\ 0 \right) \|_1
        \) & -15.0 \\
        Joint vel limits & \( \left\| \min\left( \max\left( \left| \dot{\mathbf{q}} \right| - \dot{\mathbf{q}}_{\text{lim}},\ 0 \right),\ 1.0 \right) \right\|_1 \) & -10.0\\
        Joint pos limits & \(\left\| \max\left( \mathbf{q} - \mathbf{q}_{\text{lim}}^{\text{upper,soft}},\ 0 \right) + \min\left( \mathbf{q} - \mathbf{q}_{\text{lim}}^{\text{lower,soft}},\ 0 \right) \right\|_1\) & -10.0 \\
        Upside down termination & \( \mathbf{1} \left[ \text{flipped termination} \right]\) & -5000 \\    
        \bottomrule
    \end{tabular}
    \vspace{0.5em}
    \label{tab:regularization_rewards}
\end{table}

\paragraph{Critic Decomposition.}
In DUSDi \citep{hu_disentangled_2024}, the authors also propose decomposing the Q function as a sum of Q values over individual factors. We do the same, but with value functions. Additionally, instead of having one value function per factor, we may have an ensemble of value functions per factor due to UCB exploration \citep{chen2017ucbexplorationqensembles}. To do so, we need to store the individual factor rewards separately, as the aggregated rewards are only required for the policy update. As a result, we do weighted aggregation over the advantage estimates:
\begin{align}
    \advantage = \lambda_{N+1} \advantage_{\text{style}} + \sum_{i=1}^{N} \lambda_i (\advantage_{i,\mu} + \lambda_{\text{UCB}} \advantage_{i,\sigma}).
\end{align}
where \( \advantage_{\text{style}} \) is the advantage of the style factor and \(\advantage_{i,\mu}, \advantage_{i,\sigma}\) are the mean and standard deviation of the advantage ensembles per factor. The policy is updated with aggregated advantage \( \advantage \).

\begin{figure}[htbp]
    \centering
    \begin{subfigure}[t]{0.48\textwidth}
        \includegraphics[width=\linewidth]{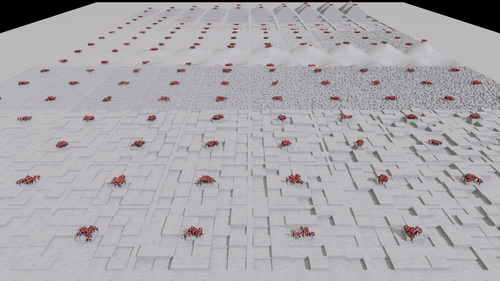}
        \caption{Game-based curriculum terrain design.}
        \label{fig:terrain_rough_curr}
    \end{subfigure}
    \hfill
    \begin{subfigure}[t]{0.48\textwidth}
        \includegraphics[width=\linewidth]{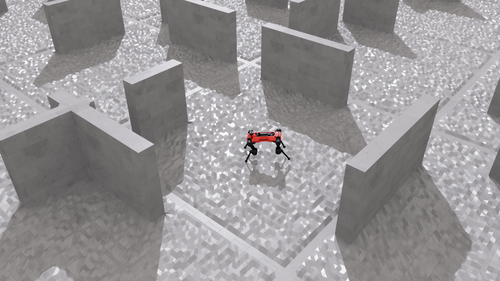}
        \caption{Rough terrain environment with random walls.}
        \label{fig:terrain_walls}
    \end{subfigure}

    \caption{\textbf{Environments used for learning  skills.} These environments are generated procedurally using the same mechanism as in~\citet{rudin_advanced_2022}. The policy receives the height-scan for perceiving the different terrains.}
    \label{fig:terrains}
\end{figure}

\paragraph{Environments.}
\label{sec:app_envs}
The simulation environments are implemented in NVIDIA Isaac Lab~\cite{mittal2023orbit}. \cref{fig:terrain_rough_curr} shows the different types of terrain used in the experiments. The environment comprises flat, randomly rough, and pyramidal sloped and stair terrains. The robot is placed at the center of these sub-terrains and no external task-specific objectives are provided from the environment. The robot needs to learn diverse skills through its intrinsic USD objectives.

Similarly to previous work~\cite{rudin_advanced_2022}, we randomize the physical properties of the robot (such as friction and base mass) and introduce external pushes for robustness. An episode end if the robot base rotates more than \SI{100}{\deg} or the duration of the episode reaches \SI{30}{\second}.

Inspired by~\citet{rudin_advanced_2022}, we design a game-inspired terrain curriculum where the robot encounters increasingly difficult sub-terrains as training progresses. Our curriculum is not driven by a task-specific reward, but rather by a task-agnostic measure of skill capability: state coverage. Since increasingly difficult terrain primarily challenges the agent's ability to explore its position state space, we use the coverage of this specific factor to control the curriculum's progression. We quantify this coverage by the total distance traversed, and adjust the terrain difficulty based on performance: if an agent travels more than \SI{10.0}{\meter}, it advances to a more difficult level, whereas if it travels less than \SI{5.0}{\meter}, it is moved to an easier one.

The environments shown in \cref{fig:terrain_box,fig:terrain_obstacle_curr,fig:terrain_walls} are used for additional experiments on discovering loco-manipulation and high-level navigation skills. These are discussed in~\cref{sec:app_additional_experiments}.

\subsection{Evaluation Metrics}
\label{app:eval_metrics}

\paragraph{Metric Score.}
The metric score, a value in \([-1, 1]\), quantifies the performance of each skill factor. The definition of the score varies by the type of factor.
\begin{itemize}
    \item For METRA factors, the score is the cosine similarity between the latent state transition \((\staterepr(\state') - \staterepr(\state))\) and the commanded skill \(\skill\). This corresponds directly to the METRA reward signal, \(\reward_{\text{METRA}}(\state, \skill, \state')\).
    \item For DIAYN factors, the score is the cosine similarity between the commanded skill \(\skill\) and the expectation of the predicted posterior, \(\mathbb{E} [\discriminator_{\phi}(\skill \mid \state)]\). Note: For a Dirichlet posterior, this value is always positive, as its support is the probability simplex.
    \item For the style factor, we directly use the scaled extrinsic reward as the metric score.
\end{itemize}
Due to these different definitions, metric scores are not comparable across different factor types and should only be compared for the same factor across different experimental runs.

\paragraph{Diversity.}
To quantify the diversity of learned behaviors, we measure the breadth of the state space that the policy can reach. We calculate this metric as follows:
\begin{enumerate}
    \item Sample a large number of skills, \(n > 10,000\), from the prior \(p(\skill)\).
    \item For each skill, execute a full rollout with the policy to collect a trajectory of states.
    \item Calculate the mean state for each of the \(n\) trajectories.
    \item Calculate the standard deviation over these \(n\) mean states.
\end{enumerate}
This final standard deviation serves as our diversity metric, where higher values indicate broader state coverage.

\subsection{Additional Results and Discussion}
\label{app:exp_details} 

In this section, we provide additional details and insights for the experiments in the main paper.

\paragraph{Mixing USD Algorithms for Diverse Factor Types.}
The results in \cref{tab:alg_mixing_results} highlight a key insight beyond the performance trade-off mentioned in the main text: single-algorithm baselines tend to over-specialize. We observe that when a method struggles with one type of factor (e.g., unbounded position), its measured success on another (e.g., bounded heading) can be inflated. Our mixed approach avoids this issue by leveraging each algorithm's strengths. This principle also guides our choice of skill dimensions, where we use a 2D skill for METRA to match the xy-plane's geometry and a 4D skill for DIAYN to represent discrete-like heading directions.

\paragraph{Investigating Reward Weighting for Conflicting Factors.}

\begin{wrapfigure}{r}{0.5\textwidth}
  \vspace{-2em}
  \centering
  \includegraphics[width=0.48\textwidth]{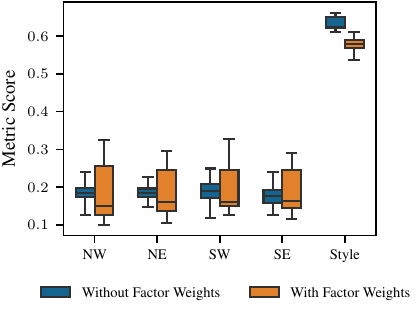}
  \caption{Effect of factor weighting on conflicting factors. Weighting does not improve the metrics} 
  \vspace{-0.5em}
  \label{fig:factor_weights_effect_conflicts}
\end{wrapfigure}
We expected factor weighting to alleviate the issue in factorized skill learning when certain factors cannot be interacted with simultaneously. To evaluate this, we factorized the position factor further into four quadrants (NE ($x>0, y>0$), SE ($x>0, y<0$), SW ($x<0, y<0$), and NW ($x<0, y>0$)). The robot cannot be in multiple quadrants simultaneously, which conflicts with the skills commanded for each factor in each quadrant. We trained all factors with METRA and without symmetry augmentation. In \cref{fig:factor_weights_effect_conflicts} we visualize the cosine similarity between the latent state transition and the commanded skills, and the scaled style reward for setups with and without weighting. Weighting does not change the performance significantly. We hypothesize that factor weights did not help because while they affect the policy's reward, the underlying USD networks (discriminators/encoders) are still trained on all collected data. This means data from trajectories where a factor’s weight was near zero is still used to train that factor's USD network, creating conflicting gradients. This suggests a valuable future direction: incorporating the factor weights into the USD network loss, such that the collected rollouts are weighted by their relevance to each factor during the network update.

\begin{figure}[htbp]
    \centering
    \begin{minipage}[t]{0.48\textwidth}
        \centering
        \includegraphics[width=\linewidth]{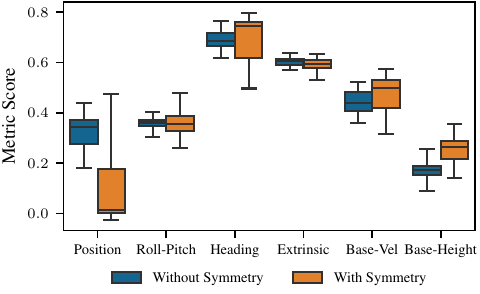}
        \caption{Effect of symmetry augmentation on skill learning metrics.}
        \label{fig:symmetry_convergence}
    \end{minipage}
    \hfill
    \begin{minipage}[t]{0.48\textwidth}
        \centering
        \includegraphics[width=\linewidth]{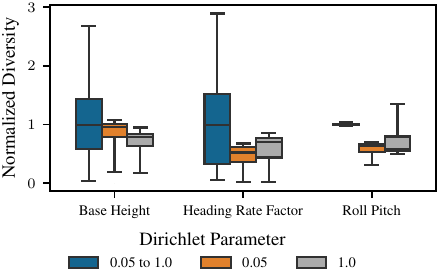}
        \caption{Effect of Dirichlet parameter on diversity on three factors.}
        \label{fig:dirichlet_param_diversity}
    \end{minipage}
\end{figure}

\paragraph{Symmetry Augmentation.}
In \cref{fig:symmetry_convergence}, we show how symmetry augmentation affects skill discovery performance across different state factors. 
METRA is used for the position factor, while DIAYN is used for the others.
Notably, symmetry augmentation reduces the encoder accuracy in METRA, while it has little effect on the discriminator accuracy in DIAYN. We hypothesize that this is because METRA relies on a more direct, directional interpretation of the skill vector, making it more sensitive to the constraints introduced by symmetry augmentation.

\paragraph{DIAYN Distribution Type.}  %
We evaluate the impact of different Dirichlet priors on skill diversity for DIAYN-trained factors. We compare three setups: a fixed high-concentration prior (\(\alpha = 0.05\),
  a fixed low-concentration prior (\(\alpha = 1.0\)), and a curriculum that linearly increases \(\alpha\) from 0.05 to 1.0 based on discriminator accuracy. 
As shown in \cref{fig:dirichlet_param_diversity}, the curriculum setup results in greater variability and often higher diversity scores for the base height and heading rate factors. This suggests that starting with sparse skill sampling helps early specialization, while gradually broadening the support encourages later diversity. In contrast, the Roll-Pitch factor appears less sensitive to the choice of prior, likely due to its lower inherent diversity. Overall, the curriculum provides a trade-off between diversity and training stability.

\subsection{Additional Tasks}
\label{sec:app_additional_experiments}

\paragraph{Loco-manipulation.}
We attempted to learn loco-manipulation skills by placing a movable box in the environment (shown in~\cref{fig:terrain_box}) and adding the box pose as an additional state factor. However, this setup alone failed to produce meaningful skills, as interactions with the box were rare and the resulting intrinsic rewards from the box factor were weak. To address this, we incorporated exploration guidance using RND~\citep{burda2018explorationrandomnetworkdistillation} and UCB~\citep{chen2017ucbexplorationqensembles}. Neither method led to significant improvements. With RND, the prediction error rapidly decreased before the agent could discover interactions with the box, providing little incentive to explore it. With UCB, the agent received persistently high intrinsic rewards due to high ensemble disagreement, caused by low-quality value estimates, without corresponding learning progress, leading to unstructured and unproductive behavior. 

\citet{strouse_learning_2022_optimistic} showed that DIAYN suffers from poor exploration. To still encourage high diversity, we can add an exploration bonus on top of the pure skill discovery reward.
A simple form of intrinsic motivation is random network distillation (RND) \citep{burda2018explorationrandomnetworkdistillation}, which encourages exploration by rewarding states that are rarely visited. 
Another method to encourage exploration is to use ensemble disagreement as an exploration reward. One way to implement this, proposed by \citet{strouse_learning_2022_optimistic}, is by defining multiple discriminators \( \discriminator_{\phi_i}(\skill \mid \state) \) and then rewarding the agent for high entropy of the mixture compared to the mean entropy.
\begin{align}
    \reward_{\text{DISDAIN}} = \entropy\left( \frac{1}{N} \sum_{i=0}^N \discriminator_{\phi_i}(\skill \mid \state)  \right) -  \frac{1}{N} \sum_{i=0}^N \entropy\left(\discriminator_{\phi_i}(\skill \mid \state)  \right)
\end{align}

Depending on the distribution, this may not be easy to implement. A simpler approach based on ensemble disagreement uses the variance of the rewards as an exploration bonus:
\begin{align}
    \reward_{\text{EXPLORE}} =  \text{Var}([\log \discriminator_{\phi_1}(\skill \mid \state), \dots, \log \discriminator_{\phi_N}(\skill \mid \state)])
\end{align}

This is similar to the method proposed by \citet{chen2017ucbexplorationqensembles}, which used a Bayesian learning approach by updating the policy based on an upper confidence bound (UCB) for the value estimate by defining an ensemble of value functions and adding a disagreement bonus. However, this method also did not help to discover meaningful loco-manipulation skills.

\paragraph{High-level Navigation.}
We investigated whether our framework could be applied hierarchically to learn complex navigation behaviors without direct supervision. For this, we used one of our pre-trained USD policies as a fixed, low-level controller that provides a library of basic skills. We then trained a high-level policy on top, again using our USD objective, which learns to sequence these skills by outputting skill vectors for the low-level policy. We evaluated this in environments with randomly placed obstacles (shown in \cref{fig:terrain_obstacle_curr,fig:terrain_walls}), applying the METRA objective on the base position factor to train the high-level policy. While the agent learned to cover the space, it struggled to avoid obstacles, even with access to a 2D planar distance scan. We hypothesize that this is due to the lack of an extrinsic signal that encourages obstacle avoidance.

We also explored using this setup to learn box-pushing behaviors. To simplify the task, the box could be moved by simple collisions. Using METRA on the box position factor, the agent learned to push the box effectively. 
However, since the low-level policy was not trained in the presence of the box, it lacked any meaningful manipulation capabilities. As a result, the agent relied on forceful collisions to move the box, an approach that is unsafe for real-world deployment.

\end{document}